\documentclass[runningheads]{llncs}

\usepackage[T1]{fontenc}

\usepackage{graphicx}

\usepackage{euscript}
\usepackage{amssymb}
\usepackage{multirow}
\usepackage[misc]{ifsym}
\usepackage[super]{nth}
\usepackage{diagbox}
\usepackage{hhline}
\usepackage{subcaption}
\usepackage[hidelinks]{hyperref}

\usepackage[table]{xcolor}
\usepackage{ulem}
\usepackage{orcidlink}

\newcommand{\boldparagraph}[1]{\paragraph{{\textbf{#1}}}}

\usepackage{xcolor,colortbl}
\usepackage{colortbl}
\usepackage{arydshln}

\usepackage{dcolumn}
\newcolumntype{C}{D{.}{.}{2.2}}

\usepackage{glossaries}

\newacronym{gdloss}{GD loss}{Generalized Dice Loss}
\newacronym{celoss}{CE loss}{Cross-Entropy Loss}

\newacronym[plural=CNNs,firstplural=Convolutional Neural Networks (CNNs)]{cnn}{CNN}{Convolutional Neural Network}
\newacronym{cca}{CCA}{connected component analysis}
\newacronym{mas}{MAS}{Multi-Atlas Segmentation}
\newacronym{mcd}{MCD}{Monte-Carlo Dropout}

\newacronym{iid}{i.i.d.}{independent and identically distributed}
\newacronym{tl}{TL}{Transfer Learning}
\newacronym{da}{DA}{Domain Adaptation}
\newacronym{dg}{DG}{Domain Generalization}
\newacronym{dr}{DR}{Domain Randomization}

\newacronym{ct}{CT}{Computed Tomography}
\newacronym{mr}{MR}{Magnetic Resonance}
\newacronym{hu}{HU}{Hounsfield Units}

\newacronym{miccai}{MICCAI}{International Conference on Medical Image Computing and Computer-Assisted Intervention}
\newacronym{mmwhs}{MMWHS}{Multi-Modality Whole Heart Segmentation}

\newacronym[plural=ROIs,firstplural=regions of interest (ROI)]{roi}{ROI}{region of interest}
\newacronym[plural=GPUs,firstplural=graphics processing units (GPU)]{gpu}{GPU}{graphics processing unit}

\newacronym{lv}{LV}{left ventricle}
\newacronym{rv}{RV}{right ventricle}
\newacronym{la}{LA}{left atrium}
\newacronym{ra}{RA}{right atrium}
\newacronym{myo}{MYO}{myocardium}
\newacronym{aa}{AA}{ascending aorta}
\newacronym{pa}{PA}{pulmonary artery}

\newacronym{dsc}{DSC}{Dice Similarity Coefficient}
\newacronym{hd}{HD}{Hausdorff Distance}
\newacronym{assd}{ASSD}{Average Symmetric Surface Distance}

\newacronym{rsc}{RSC}{representation self-challenging}

\newacronym{care2024}{CARE2024}{Comprehensive Analysis \& computing of REal-world medical images}
\newacronym{rc}{RC}{RandConv-based augmentation}

\newcommand{\lossbase}{L}
\newcommand{\lossdice}{\lossbase_{\text{GD}}}

\newcommand{\imagelettersmall}{x}
\newcommand{\groundtruthlettersmall}{y}

\newcommand{\image}{\mathbf{\imagelettersmall}}

\newcommand{\groundtruth}{\mathbf{\groundtruthlettersmall}}

\newcommand{\prediction}{\mathbf{\hat{\groundtruthlettersmall}}}

\newcommand{\imagect}{\image^{\text{CT}}}
\newcommand{\imagemr}{\image^{\text{MR}}}
\newcommand{\groundtruthct}{\groundtruth^{\text{CT}}}
\newcommand{\groundtruthmr}{\groundtruth^{\text{MR}}}
\newcommand{\predictionct}{\prediction^{\text{CT}}}
\newcommand{\predictionmr}{\prediction^{\text{MR}}}

\newcommand{\rnanetletter}{g}
\newcommand{\rnanet}{\rnanetletter}

\newcommand{\rnaalpha}{\alpha}

\newcommand{\weightsletter}{\theta}

\newcommand{\model}{\mathcal{M}}

\usepackage{color}

\begin{document}
\title{Augmentation-based Domain Generalization \\ and Joint Training from Multiple Source Domains \\ for Whole Heart Segmentation}

\titlerunning{Domain Generalization and Joint Training for Whole-Heart Segmentation}

\author{
Franz Thaler\inst{1,2}\orcidID{0000-0002-6589-6560} \and
Darko \v{S}tern\inst{3}\orcidID{0000-0003-3449-5497} \and
\\
Gernot Plank\inst{1}\orcidID{0000-0002-7380-6908} \and
Martin Urschler\inst{4}\orcidID{0000-0001-5792-3971}
}

\authorrunning{F. Thaler et al.}

\institute{
Gottfried Schatz Research Center: Medical Physics and Biophysics, \\Medical University of Graz, Graz, Austria \and
Institute of Computer Graphics and Vision, Graz University of Technology, \\Graz, Austria \and
AVL List GmbH, Graz, Austria \and
Institute for Medical Informatics, Statistics and Documentation, \\Medical University of Graz, Graz, Austria\\
}

\maketitle 

\begin{abstract}
As the leading cause of death worldwide, cardiovascular diseases motivate the development of more sophisticated methods to analyze the heart and its substructures from medical images like Computed Tomography (CT) and Magnetic Resonance (MR).
Semantic segmentations of important cardiac structures that represent the whole heart are useful to assess patient-specific cardiac morphology and pathology.
Furthermore, accurate semantic segmentations can be used to generate cardiac digital twin models which allows e.g. electrophysiological simulation and personalized therapy planning.
Even though deep learning-based methods for medical image segmentation achieved great advancements over the last decade, retaining good performance under domain shift – i.e. when training and test data are sampled from different data distributions – remains challenging.
In order to perform well on domains known at training-time, we employ a (1) balanced joint training approach that utilizes CT and MR data in equal amounts from different source domains.
Further, aiming to alleviate domain shift towards domains only encountered at test-time, we rely on (2) strong intensity and spatial augmentation techniques to greatly diversify the available training data.
Our proposed whole heart segmentation method, a 5-fold ensemble with our contributions, achieves the best performance for MR data overall and a performance similar to the best performance for CT data when compared to a model trained solely on CT.
With $93.33$\% DSC and $0.8388$ mm ASSD for CT and $89.30$\% DSC and $1.2411$ mm ASSD for MR data, our method demonstrates great potential to efficiently obtain accurate semantic segmentations from which patient-specific cardiac twin models can be generated.
\end{abstract}

\section{Introduction}

Cardiovascular diseases are the leading cause of death worldwide~\cite{vaduganathan2022global}, which motivates more in-depth analysis of the heart and its substructures in clinical practice.
Whole heart segmentation (WHS) obtained from medical images like \gls{ct} or \gls{mr} is useful to assess morphological and pathological changes of the patient-specific anatomy.
Moreover, WHS is crucial for the generation of anatomically accurate and patient-specific cardiac twin models of e.g. human electrophysiology~\cite{gillette2021framework}, which enable downstream tasks like electrophysiological simulation in order to allow personalized therapy planning \cite{Campos2022}.
However, manually segmenting the whole heart is a very time-consuming and tedious process that requires an expert with extensive anatomical knowledge.
Acquiring a manual segmentation of the whole heart takes about 6 to 10 hours for a single subject~\cite{Zhuang2016-dz}, a task that would benefit tremendously by automation.
Deep learning based neural networks achieved substantial advancements over the last decade in a multitude of image analysis tasks like the detection of diseases in medical images~\cite{Esteva2017-ci,McKinney2020-js}, where they even performed better than humans.
Nevertheless, two challenges arise in the context of machine learning in order to achieve such excellent results: First, they typically require large amounts of training data and second, they rely on the assumption that training and test data are \gls{iid}.
In the presence of domain shift, i.e. when training and test data are not sampled from the same distribution, machine learning algorithms are shown to underperform, resulting in higher test errors~\cite{Ben-David2006-ck,Torralba2011-gm} and potentially even in complete model failure~\cite{AlBadawy2018-gi,Pooch2020-vz}.
Several factors contribute to domain shift and may impact the model's performance, among which are differences in intensity values, contrast and morphology.
Intensity differences are introduced by the imaging modality, e.g. \gls{ct} or \gls{mr}, imaging protocol, use of contrast agents, or the scanner model, see e.g.~\cite{Zhuang2019MvMM}.
Meanwhile, morphological differences may result from patient-specific diseases or from inherent differences in the local population.

\begin{figure*}[t]
\includegraphics[width=\textwidth]{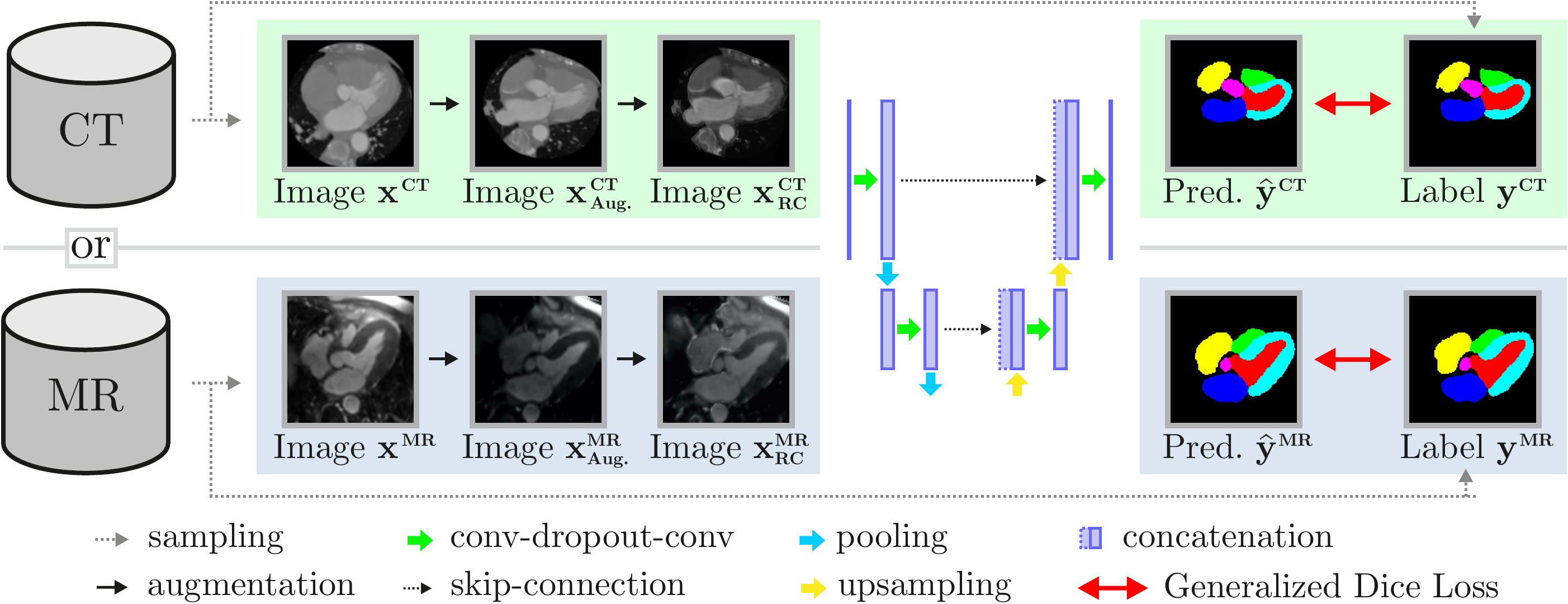}
\caption{
Overview of the proposed method, where we employ a balanced joint training strategy as well as strong intensity and spatial augmentation techniques.
During training, we sample a batch consisting of a \gls{ct} and an \gls{mr} scan, i.e. $\image^{\text{CT}}$ and $\image^{\text{MR}}$, from the available training data, which is then augmented using spatial and intensity augmentation techniques yielding $\image_{\text{Aug.}}^{\text{CT}}$ and $\image_{\text{Aug.}}^{\text{MR}}$.
Next, a RandConv-based augmentation technique is applied to further diversify the data resulting in $\image_{\text{RC}}^{\text{CT}}$ and $\image_{\text{RC}}^{\text{MR}}$.
In-line with the DG philosophy, the proposed model uses the same weights for \gls{ct} and \gls{mr} data and each scan is processed independently by the model.
While joint training aims to maximize the performance on domains known at training-time, the augmentation techniques are employed to better bridge the domain gap to previously unknown domains.
}
\label{fig:overview}
\end{figure*}

In literature, several transfer learning strategies have been proposed to mitigate domain shift of which two prominent directions are \gls{da} and \gls{dg} approaches.
While \gls{da} typically aims to achieve a good performance on one or several target domains that are known in advance~\cite{Guan2022-vr,Kumari2024-kl}, \gls{dg} aims to generalize to new target domains that are unknown during training~\cite{Zhou2023-jk,Wang2023-vz,GAO2023BayeSeg}.
In this work, we aim for both, (1) retaining a good performance on data from domains already seen during training, which is more in-line with \gls{da}, as well as (2) achieving good results on data from new domains that were previously unseen, which follows the \gls{dg} philosophy.
First, we alleviate the effect of domain shift among previously known domains by applying a balanced joint training paradigm from multiple source domains that relies on conflating the feature representation of different source data distributions.
Second, we address domain shift towards previously unknown domains by employing strong intensity and spatial augmentation techniques in order to greatly diversify the observed feature representations during training and to increase robustness.
Our method is a contribution to the WHS++ track of the CARE2024 Challenge\footnote{CARE2024 Challenge website: \url{http://www.zmic.org.cn/care_2024/}, last accessed in September, 2024} and externally evaluated in the context of this challenge.

\begin{figure*}[t]
\includegraphics[width=\textwidth]{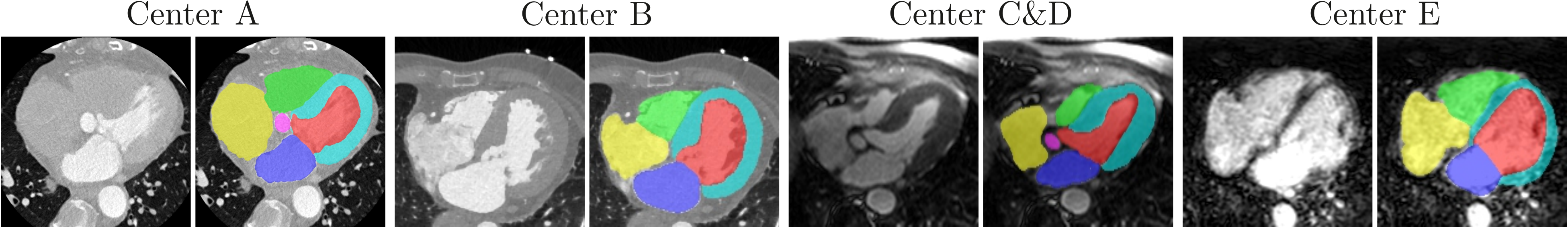}
\caption{
Exemplary training images and ground truth segmentations from different centers.
Differences in intensity values, contrast and artifacts can be identified when comparing the \gls{ct} scans (Center A and B), the \gls{mr} scans (Center C\&D and E) and most prominently, when comparing \gls{ct} and \gls{mr} to one another.
}
\label{fig:ct_mr_comparison}
\end{figure*}

\section{Method}

In this work we employ a modality-agnostic 3D \gls{cnn} that semantically segments \gls{ct} or \gls{mr} data from different domains in order to obtain accurate segmentations of seven important cardiac structures that represent the whole heart.
While the training data is comprised of labelled cardiac \gls{ct} and \gls{mr} scans from overall five different medical centers, the test data consists of medical images from previously known as well as unknown domains.
Aiming to alleviate the effect of domain shift for both of these cases, we propose a method for WHS that employs (1) a balanced joint training approach from multiple source domains, as well as (2) strong intensity and spatial augmentation techniques that diversify training data to better bridge the domain gap towards previously unknown domains, see Fig.~\ref{fig:overview}.

\boldparagraph{Joint Training from Multiple Source Domains:}

Due to differences in image acquisition, it is well-known that \gls{ct} and \gls{mr} data is visually very different in terms of intensity distribution, intensity ranges and contrast even when visualizing the same anatomy, see Fig.~\ref{fig:ct_mr_comparison}.
Moreover, differences in signal-to-noise ratio as well as the types of acquisition artifacts that might be present in the captured data, are all factors that contribute to domain shift.
However, while \gls{ct} data is relatively self-similar as intensity ranges are well-defined via the Hounsfield unit scale, \gls{mr} data tends to be less consistent to itself when obtained under different conditions as it can only be scaled relatively.
That being said, we identified the main source of domain shift as the imaging modality, i.e. \gls{ct} respectively \gls{mr}, rather than the differences that exist \textit{within} one modality.
Due to GPU memory limitations when training a \gls{cnn} on high resolution data in 3D, we decided to consider \gls{ct} and \gls{mr} data as two separate source domains that are provided in a balanced manner in the batch dimension for each training step of the network.
By doing so, we implicitly treat data of the \textit{same} modality that was acquired at different centers as internal variation within that modality by randomly sampling from the respective available training dataset.
Specifically, training data $\mathcal{D}_{tr}$ encompasses data obtained from the different centers $A$ to $E$, which we group into a set that comprises \gls{ct} data as $\mathcal{D}_{tr}^{\text{CT}} = \{ \mathcal{D}_{tr}^{A}, \mathcal{D}_{tr}^{B} \}$ and a set of \gls{mr} data as $\mathcal{D}_{tr}^{\text{MR}} = \{ \mathcal{D}_{tr}^{C}, \mathcal{D}_{tr}^{D}, \mathcal{D}_{tr}^{E} \}$.
For each iteration during training, we randomly sample one 3D image from the set of available \gls{ct} and \gls{mr} images in order to obtain $\imagect \sim \mathcal{D}_{tr}^{\text{CT}}$ and $\imagemr \sim \mathcal{D}_{tr}^{\text{MR}}$, respectively. 
These sampled images are combined in a batch, thus establishing a batch size of two.
Defining the \gls{cnn} model as $\model (\cdot)$ with trainable weights $\weightsletter$ allows expressing the prediction obtained for image $\image$ as $\prediction = \model (\image; \weightsletter)$.
Lastly, after defining the corresponding ground truth segmentation as $\groundtruth$ and the generalized Dice loss function as $\lossdice$, we can formulate our optimization target as:
\begin{equation}
    \lossbase = \frac{1}{2} \left( \lossdice(\groundtruthct, \predictionct) + \lossdice(\groundtruthmr, \predictionmr) \right).
\end{equation}

\boldparagraph{Augmentation-based Generalization:}

Before any image $\image$ is processed by the \gls{cnn} model $\model (\cdot)$, we perform preprocessing as described in Section~\ref{sec:implementation_details} and also apply our data augmentation pipeline that aims to diversify training data.
Specifically, aiming to account for differences in orientation, size and morphology of the heart in different local cohorts, we apply spatial augmentation in terms of translation, rotation, scaling and elastic deformation.
The next step in order to reduce domain shift is intensity augmentation, where we aim to compensate for visual differences like intensity values, ranges, contrast and noise that are expected to be present in the different domains.
First, we apply more conventional global intensity augmentation by employing a randomly sampled intensity shift and scale parameter, before we modify intensities per label by using an additional shift and scale parameter.
After this, we use a RandConv-based augmentation technique, which augments intensity values as such that domain-specific features are likely to be removed without corrupting the overall shape information of the anatomy~\cite{Xu2020-vo,Ouyang2023-xf,Choi2023-qy}.
In this work, we employ RandConv as a shallow multi-layer network defined as $\rnanet(\cdot)$.
The weights of $\rnanet(\cdot)$ are randomly initialized whenever a new image needs to be augmented and consequently, are never trained.
Due to the large variation of the immediate output of $\rnanet(\image)$, we perform linear interpolation with the input image $\image$ with a blending factor $\rnaalpha$ to partially reduce the augmentation effect~\cite{Hendrycks2019-ls,Xu2020-vo,Ouyang2023-xf}.
Lastly, to prevent exploding gradients during training, RandConv augmented images need to be re-normalized for which we employ the Frobenius norm $|| \cdot ||_{F}$, i.e.:
\begin{equation}
    \text{norm} (\image, \rnanet) = \frac
    {\rnaalpha \cdot \rnanet(\image) + (1 - \rnaalpha) \cdot \image}
    {|| \rnaalpha \cdot \rnanet(\image) + (1 - \rnaalpha) \cdot \image ||_{F}}
    \cdot || \image ||_{F}.
\end{equation}

\section{Experimental Setup}

\subsection{Dataset}

The dataset used in this work is part of the CARE2024 Challenge and provided for the WHS++ track.
Overall, the WHS++ dataset consists of 104 \gls{ct} and 102 \gls{mr} scans obtained from six medical centers.
The dataset is separated into a training, validation and test set with 86, 50 and 70 scans, respectively.
The exact numbers on how many images are available per center as well as their distribution among the different sets are given in Table~\ref{tab:dataset}.
Ground truth annotations are available for seven anatomical structures and follow the definition of the MMWHS dataset from the MMWHS Challenge held in 2017~\cite{Zhuang2019-uj}.
The labels represent the blood cavity of the left ventricle, right ventricle, left atrium and right atrium, the myocardium that surrounds the left ventricular blood cavity, as well as the ascending aorta and the pulmonary artery.

\begin{table*}
\centering
\caption{
The number of subjects per modality and per center included in the training, validation and test set of the WHS++ track of the CARE2024 Challenge.
Center C\&D is treated as one center by the organizers and separate numbers for C and D were not disclosed.
}
\begin{tabular}{
p{0.2\textwidth}
| >{\centering}p{0.1\textwidth}
| >{\centering}p{0.1\textwidth}
| >{\centering}p{0.1\textwidth}
| >{\centering}p{0.1\textwidth}
| >{\centering\arraybackslash}p{0.1\textwidth}
}

Modality & \multicolumn{2}{c|}{\textbf{CT}} & \multicolumn{3}{c}{\textbf{MR}} \\
Center & A & B & C\&D & E & F \\
\hline

Training Set & 20 & 20 & 20 & 26 & -- \\  
Validation Set & 20 & 10 & 20 & -- & -- \\
Test Set & 20 & 14 & 20 & -- & 16 \\

\end{tabular}

\label{tab:dataset}
\end{table*}

\subsection{Implementation Details}
\label{sec:implementation_details}

All image data is resampled to have an isotropic spacing of $1.5 \text{ mm}$ in 3D before being processed by the \gls{cnn}.
In order to reduce the GPU memory requirements during training, training data was resampled to a tight cube with an image size of $128 \times 128 \times 128$ voxel around the heart, for which we computed the center position as the center of the foreground labels in the ground truth segmentation.
Even though the approximated center position of the heart cannot be computed for test data using the same logic, it is safe to assume that the heart is located roughly in the center of the image for this type of data.
Moreover, since GPU memory requirements are reduced at test-time, we resampled images from the validation and test set to an image size of $192 \times 192 \times 192$ voxels around the center point of the whole image.

In this work, we employ the data augmentation toolkit of~\cite{Payer2018,payer2019integrating}.
For spatial augmentation we employ translation with up to $\pm 20$ voxel, rotation with up to $\pm 0.35$ radians, scaling in the range of $[0.8, 1.2]$ and elastic deformation which uses eight grid nodes per dimension and deformation values sampled from $\pm 15$ voxels.
\gls{ct} data is expected to follow the Hounsfield unit scale and is normalized by division with 2048 before clipping the data to $[-1, 1]$.
\gls{mr} data is normalized robustly by linearly normalizing the \nth{10} and \nth{90} percentile to $[-1, 1]$.
Any intensity augmentation is performed after normalization.
Random intensity shift parameters are sampled from $\pm 0.2$, while intensity scaling employs a factor sampled from $[0.8, 1.2]$ for \gls{ct} and $[0.6, 1.4]$ for \gls{mr}, respectively.
We use the Hounsfield unit scale to identify unknown test images as either \gls{ct} or \gls{mr} scans in order to apply the respective normalization procedure.
Furthermore, we post-process predictions for images of the validation and test set by removing all components that are disconnected from the largest connected component in 3D per label as well as when all foreground labels are considered as one.

The network architecture employed in this work is similar to the U-Net~\cite{Ronneberger2015-ih} and consists of a contracting and an expanding path with skip-connections and five levels.
Two convolutions with $64$ filters and an intermediate dropout layer~\cite{Srivastava2014-rv} are used at each level and the dropout rate is set to $0.1$.
Leaky ReLU is used as an activation function with alpha $0.1$ after each convolution layer excluding the last one.
Further, average pooling and linear upsampling layers are used and ''same'' padding is employed throughout the network.
As optimizer we use Adam~\cite{Kingma2015-fb} and a learning rate of $5e^{-5}$ and convolution kernels are initialized using He initializer~\cite{He2015-hz}.
Lastly, we use temporal ensembling of network weights as proposed in~\cite{Laine2017-zh}.
The RandConv-based shallow multi-layer network $\rnanet(\cdot)$ employs four convolution layers with two channels.
The kernel size is randomly set to either one or three whenever the network is re-initialized and weights are sampled from a normal distribution $\mathcal{N}(0, 1)$.
Leaky ReLU is used as intermediate activation function with an alpha parameter of 0.1 and the blending factor $\rnaalpha$ is sampled uniformly to be within 0 and 1. 
For our final submission, we trained five independent models which we treat as an ensemble when generating predictions for images of the validation and test set, where we average the individual predictions.
Training of a single model lasted roughly 30~hours, while generating a prediction from the ensemble after loading takes roughly 11~seconds per subject on an NVIDIA GeForce RTX 3090.

\begin{table*}
\centering
\caption{
Quantitative results on the validation set of the WHS++ track of the CARE2024 Challenge. 
Assessed are the Dice Similarity Coefficient (DSC) in percent, Hausdorff Distance (HD) in mm and Average Symmetric Surface Distance (ASSD) in mm as obtained through the submission system for \gls{ct} and \gls{mr} data separately.
An ablation is provided for the joint training (JT) strategy as well as the RandConv-based augmentation (RC) technique.
Specifically, \gls{ct}-only and \gls{mr}-only respectively refer to a single model trained exclusively on \gls{ct} or \gls{mr} data, while the joint model was trained using both, \gls{ct} and \gls{mr} data in a balanced manner.
Our final submission (5-fold Ensemble) is an ensemble of five independently trained joint models with RandConv-based augmentation.
The best score per metric is shown in bold.
}
\begin{tabular}{ l | c | c | c | c | c | c | c | c }

\multirow{2}{*}{\textbf{Method}} & \multirow{2}{*}{\textbf{JT}} & \multirow{2}{*}{\textbf{RC}} & \multicolumn{3}{c|}{\textbf{CT}} & \multicolumn{3}{c}{\textbf{MR}} \\
& & & DSC ($\uparrow$) & HD ($\downarrow$) & ASSD ($\downarrow$) & DSC ($\uparrow$) & HD ($\downarrow$) & ASSD ($\downarrow$) \\
\hline

CT-only &  &  & \textbf{93.77} & 13.8484 & \textbf{0.7699} & 60.86 & 45.0989 & 7.4810 \\
MR-only &  &  & 13.37 & 109.0493 & 37.9332 & 87.77 & 21.4554 & 1.4598 \\
Joint Model & \checkmark &  & 93.61 & 14.3288 & 0.7964 & 88.08 & 21.1851 & 1.4581 \\

\hline

CT-only &  & \checkmark & 93.28 & \textbf{13.6464} & 0.8395 & 81.02 & 33.2132 & 3.2741 \\
MR-only &  & \checkmark & 86.31 & 16.6588 & 1.5774 & 88.69 & 17.5594 & 1.3252 \\
Joint Model & \checkmark & \checkmark & 93.00 & 14.0950 & 0.8779 & 88.92 & 16.7884 & 1.2918 \\

\hline

5-fold Ensemble & \checkmark & \checkmark & 93.33 & 13.9054 & 0.8388 & \textbf{89.30} & \textbf{16.1008} & \textbf{1.2411} \\

\end{tabular}

\label{tab:quantitative_results}
\end{table*}

\begin{figure*}[t]
\includegraphics[width=\textwidth]{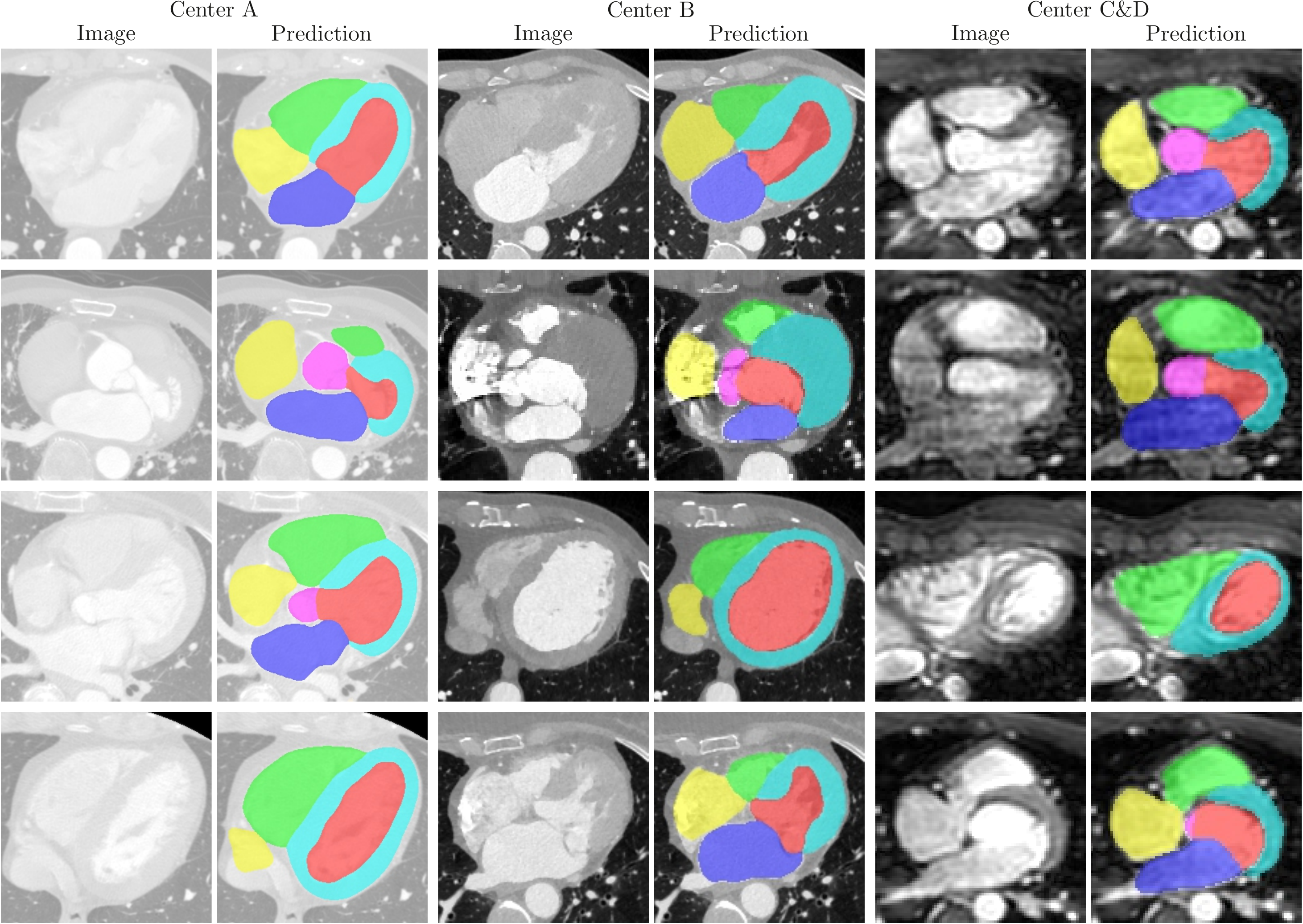}
\caption{
Qualitative results on the validation set of the WHS++ track of the CARE2024 Challenge. 
Shown are pairs of corresponding images and predictions for subjects of center A (cols. 1-2), center B (cols. 3-4) and center C\&D (cols. 5-6).
The colors correspond to the left ventricle (red), right ventricle (green), left atrium (blue), right atrium (yellow), myocardium (cyan), aorta (magenta) and pulmonary artery (white).
}
\label{fig:qualitative_results}
\end{figure*}

\section{Results and Discussion}

The evaluation conducted in this work is based on the validation set for which quantitative scores have been obtained through the submission system as provided by the organizers of the CARE2024 Challenge.
The quantitative evaluation includes separate scores for \gls{ct} and \gls{mr} data and assesses three metrics, namely, \gls{dsc} in percent, \gls{hd} in mm and \gls{assd} in mm.
Since ground truth labels are not publicly available for the validation set, the qualitative evaluation of our method is performed by visually inspecting the predictions and comparing them to the original image information.

Quantitative results are provided in Table~\ref{tab:quantitative_results}, where we compare the performance of several ablation experiments to one another.
Namely, we perform an ablation of the joint training strategy, where we show results obtained from a single model trained using exclusively \gls{ct} or \gls{mr} data to which we respectively refer to as \gls{ct}-only and \gls{mr}-only.
Further, the joint model represents a single model trained using both \gls{ct} and \gls{mr} data in a balanced manner.
For each of these, we also show ablation results when using the \gls{rc} technique.
While the \gls{ct}-only and \gls{mr}-only model without \gls{rc} achieve good results for in-domain data, i.e. data from the same domain as used during training, they are both unable to yield useful predictions for out-of-domain data, i.e. data from the respective other domain.
As expected, the joint model achieves good results for both \gls{ct} and \gls{mr} data even without \gls{rc}.
However, when the \gls{ct}-only and \gls{mr}-only model is trained with \gls{rc}, their in-domain performance remains similar, while their out-of-domain performance greatly improves compared to the \gls{ct}-only and \gls{mr}-only model without \gls{rc}.
The joint model with \gls{rc} achieved an additional improvement compared to the joint model without \gls{rc} when predicting \gls{mr} data, however, only the \gls{hd} improved when predicting \gls{ct} data.
Interestingly, the \gls{ct}-only model performed slightly better on \gls{ct} data than the joint model, while the joint model achieved better results on \gls{mr} data compared to the \gls{mr}-only model.
These results imply that a model trained and evaluated on \gls{mr} benefits from additional training data even when it is \gls{ct} data, while this appears not to be the case in the other direction.
One possible reason for this observation that is also in-line with our results is, that generating semantic segmentations of the whole-heart is more difficult for \gls{mr} than for \gls{ct} due to data being less consistent and often more burdened by noise or artifacts.
Further, this might also indicate that the available ground truth segmentations are more accurate for \gls{ct} data than for \gls{mr} data, which could explain why adding \gls{mr} data to the training set negatively impacts the models performance on \gls{ct} data.
Lastly, when comparing these methods to our final method, a 5-fold ensemble for which we averaged predictions of five independently trained joint models with \gls{rc}, it becomes apparent that the ensemble is able to further improve the results.
Specifically, our 5-fold ensemble achieved the best performance for each metric for \gls{mr} data and a performance that is very close to the best results for \gls{ct} data from the CT-only model.

Qualitative results on the validation set are shown in Fig.~\ref{fig:qualitative_results}, where we provide pairs of corresponding images and predictions for subjects of center A (cols. 1-2), center B (cols. 3-4) and center C\&D (cols. 5-6).
Labels represent the left ventricle (red), right ventricle (green), left atrium (blue), right atrium (yellow), myocardium (cyan), aorta (magenta) and pulmonary artery (white).
Overall, the semantic segmentation results are very convincing and represent the subject's anatomy well.
However, some minor mistakes can be identified, e.g., undersegmentation of the right atrium (row 4, cols. 1-2 and row 4, cols. 3-4), or undersegmentation of the right ventricle (row 3, cols. 5-6).
Nevertheless, most regions have been segmented very accurately despite e.g. the presence of artifacts (row 2, cols. 3-4 and row 2, cols. 5-6).
Moreover, our method was able to accurately distinguish between visually very similar structures, most prominently the compact myocardium (cyan) and the papillary muscles which have been annotated as being part of the left ventricle blood pool (red) in the ground truth segmentations (row 1, cols 3-4 and row 4, cols 3-4).

\section{Conclusion}

In this work we presented a domain-agnostic \gls{cnn}-based method to semantically segment seven important cardiac structures that represent the whole heart.
Our method is designed to achieve a good performance on domains that are already known at training-time similarly to approaches in \gls{da}, while also aiming for good results on domains that are unknown during training inspired by \gls{dg}.
To this end, our \gls{cnn} relies on (1) balanced joint training from multiple source domains, which aims to conflate the feature representations of the available training domains, as well as (2) strong intensity and spatial augmentation techniques to alleviate domain shift towards unknown domains encountered at test-time.
We evaluate the joint training paradigm by comparing the results to models that have only been trained and evaluated on \gls{ct} and \gls{mr} data, respectively.
Moreover, we also assess model performance with and without using the RandConv-based augmentation technique.
Furthermore, our final method consisting of a 5-fold ensemble of models trained with random weight initialization and data augmentation parameters achieved the best performance for \gls{mr} data and a performance similar to the best performance when predicting \gls{ct} data.
Our 5-fold ensemble achieved a \gls{dsc} of $93.33$\% for \gls{ct} and $89.30$\% for \gls{mr} data, which underlines its good performance on both modalities.
With this, our method shows great potential to speed up the generation of semantic segmentations and in turn, the generation of patient-specific cardiac twin models which can be used e.g. for electrophysiological simulation and personalized therapy planning.

\begin{credits}
\subsubsection{\ackname}
This research was funded in whole or in part by the Austrian Science Fund (FWF) 10.55776/PAT1748423 and also by the CardioTwin grant I6540 from the Austrian Science Fund (FWF).

\subsubsection{\discintname}
The authors have no competing interests to declare that are relevant to the content of this article.

\end{credits}

\bibliographystyle{splncs04}
\bibliography{references}

\begin{thebibliography}{10}
\providecommand{\url}[1]{\texttt{#1}}
\providecommand{\urlprefix}{URL }
\providecommand{\doi}[1]{https://doi.org/#1}

\bibitem{AlBadawy2018-gi}
AlBadawy, E.A., Saha, A., Mazurowski, M.A.: Deep learning for segmentation of brain tumors: Impact of cross-institutional training and testing. Medical Physics  \textbf{45}(3),  1150--1158 (2018). \doi{10.1002/mp.12752}

\bibitem{Ben-David2006-ck}
Ben-David, S., Blitzer, J., Crammer, K., Pereira, F.C.: Analysis of representations for domain adaptation. Advances in Neural Information Processing Systems  \textbf{19},  137--144 (2006). \doi{10.7551/mitpress/7503.003.0022}

\bibitem{Campos2022}
Campos, F.O., Neic, A., {Mendonca Costa}, C., Whitaker, J., O'Neill, M., Razavi, R., Rinaldi, C.A., Scherr, D., Niederer, S.A., Plank, G., et~al.: {An Automated Near-Real Time Computational Method for Induction and Treatment of Scar-related Ventricular Tachycardias.} Medical Image Analysis  \textbf{80},  102483 (aug 2022). \doi{10.1016/j.media.2022.102483}

\bibitem{Choi2023-qy}
Choi, S., Das, D., Choi, S., Yang, S., Park, H., Yun, S.: Progressive random convolutions for single domain generalization. In: {IEEE/CVF} Conference on Computer Vision and Pattern Recognition ({CVPR}). pp. 10312--10322 (2023). \doi{10.1109/CVPR52729.2023.00994}

\bibitem{Esteva2017-ci}
Esteva, A., Kuprel, B., Novoa, R.A., Ko, J., Swetter, S.M., Blau, H.M., Thrun, S.: Dermatologist-level classification of skin cancer with deep neural networks. Nature  \textbf{542}(7639),  115--118 (2017). \doi{10.1038/nature21056}

\bibitem{GAO2023BayeSeg}
Gao, S., Zhou, H., Gao, Y., Zhuang, X.: {BayeSeg: Bayesian modeling for medical image segmentation with interpretable generalizability}. Medical Image Analysis  \textbf{89},  102889 (2023)

\bibitem{gillette2021framework}
Gillette, K., Gsell, M.A., Prassl, A.J., Karabelas, E., Reiter, U., Reiter, G., Grandits, T., Payer, C., {\v{S}}tern, D., Urschler, M., et~al.: {A Framework for the Generation of Digital Twins of Cardiac Electrophysiology from Clinical 12-leads ECGs}. Medical Image Analysis  \textbf{71},  102080 (2021). \doi{10.1016/j.media.2021.102080}

\bibitem{Guan2022-vr}
Guan, H., Liu, M.: Domain adaptation for medical image analysis: A survey. IEEE Transactions on Biomedical Engineering  \textbf{69}(3),  1173--1185 (2022). \doi{10.1109/TBME.2021.3117407}

\bibitem{He2015-hz}
He, K., Zhang, X., Ren, S., Sun, J.: Delving deep into rectifiers: Surpassing human-level performance on {ImageNet} classification. In: {IEEE} International Conference on Computer Vision ({ICCV}). pp. 1026--1034 (2015)

\bibitem{Hendrycks2019-ls}
Hendrycks, D., Mu, N., Cubuk, E.D., Zoph, B., Gilmer, J., Lakshminarayanan, B.: {AugMix: A Simple Data Processing Method to Improve Robustness and Uncertainty}. arXiv preprint arXiv:1912.02781  (2019)

\bibitem{Kingma2015-fb}
Kingma, D.P., Ba, J.: Adam: A method for stochastic optimization. In: International Conference on Learning Representations ({ICLR}) (2015)

\bibitem{Kumari2024-kl}
Kumari, S., Singh, P.: Deep learning for unsupervised domain adaptation in medical imaging: Recent advancements and future perspectives. Computers in Biology and Medicine  \textbf{170}(3),  107912 (2024). \doi{10.1016/j.compbiomed.2023.107912}

\bibitem{Laine2017-zh}
Laine, S., Aila, T.: Temporal ensembling for {Semi-Supervised} learning. In: International Conference on Learning Representations ({ICLR}) (2017)

\bibitem{McKinney2020-js}
McKinney, S.M., Sieniek, M., Godbole, V., Godwin, J., Antropova, N., Ashrafian, H., Back, T., Chesus, M., Corrado, G.S., Darzi, A., Etemadi, M., Garcia-Vicente, F., Gilbert, F.J., Halling-Brown, M., Hassabis, D., Jansen, S., Karthikesalingam, A., Kelly, C.J., King, D., Ledsam, J.R., Melnick, D., Mostofi, H., Peng, L., Reicher, J.J., Romera-Paredes, B., Sidebottom, R., Suleyman, M., Tse, D., Young, K.C., De~Fauw, J., Shetty, S.: International evaluation of an {AI} system for breast cancer screening. Nature  \textbf{577}(7788),  89--94 (2020). \doi{10.1038/s41586-019-1799-6}

\bibitem{Ouyang2023-xf}
Ouyang, C., Chen, C., Li, S., Li, Z., Qin, C., Bai, W., Rueckert, D.: {Causality-Inspired} {Single-Source} domain generalization for medical image segmentation. IEEE Transactions on Medical Imaging  \textbf{42}(4),  1095--1106 (2023). \doi{10.1109/TMI.2022.3224067}

\bibitem{Payer2018}
Payer, C., {\v{S}}tern, D., Bischof, H., Urschler, M.: {Multi-label Whole Heart Segmentation using CNNs and Anatomical Label Configurations}. In: Statistical Atlases and Computational Models of the Heart. ACDC and MMWHS Challenges. STACOM 2017. Lecture Notes in Computer Science(). vol. 10663, pp. 190--198. Springer (2018). \doi{10.1007/978-3-319-75541-0_20}

\bibitem{payer2019integrating}
Payer, C., {\v{S}}tern, D., Bischof, H., Urschler, M.: {Integrating spatial configuration into heatmap regression based CNNs for landmark localization}. Medical Image Analysis  \textbf{54},  207--219 (2019). \doi{10.1016/j.media.2017.09.003}

\bibitem{Pooch2020-vz}
Pooch, E.H.P., Ballester, P., Barros, R.C.: {Can We Trust Deep Learning Based Diagnosis? The Impact of Domain Shift in Chest Radiograph Classification}. In: Thoracic Image Analysis. {TIA} 2020 {MICCAI} Workshop. pp. 74--83 (2020). \doi{10.1007/978-3-030-62469-9\_7}

\bibitem{Ronneberger2015-ih}
Ronneberger, O., Fischer, P., Brox, T.: {U-Net}: Convolutional networks for biomedical image segmentation. In: Medical Image Computing and {Computer-Assisted} Intervention (MICCAI). pp. 234--241 (2015). \doi{10.1007/978-3-319-24574-4\_28}

\bibitem{Srivastava2014-rv}
Srivastava, N., Hinton, G., Krizhevsky, A., Sutskever, I., Salakhutdinov, R.: Dropout: A simple way to prevent neural networks from overfitting. J. Mach. Learn. Res.  \textbf{15}(1),  1929--1958 (2014)

\bibitem{Torralba2011-gm}
Torralba, A., Efros, A.A.: Unbiased look at dataset bias. In: {IEEE} Conference on Computer Vision and Pattern Recognition ({CVPR}). pp. 1521--1528 (2011). \doi{10.1109/CVPR.2011.5995347}

\bibitem{vaduganathan2022global}
Vaduganathan, M., Mensah, G.A., Turco, J.V., Fuster, V., Roth, G.A.: The global burden of cardiovascular diseases and risk: a compass for future health. Journal of the American College of Cardiology  \textbf{80}(25),  2361--2371 (2022)

\bibitem{Wang2023-vz}
Wang, J., Lan, C., Liu, C., Ouyang, Y., Qin, T., Lu, W., Chen, Y., Zeng, W., Yu, P.S.: Generalizing to unseen domains: A survey on domain generalization. IEEE Transactions on Knowledge and Data Engineering  \textbf{35}(8),  8052--8072 (2023). \doi{10.1109/TKDE.2022.3178128}

\bibitem{Xu2020-vo}
Xu, Z., Liu, D., Yang, J., Raffel, C., Niethammer, M.: Robust and generalizable visual representation learning via random convolutions. arXiv preprint arXiv:2007.13003  (2020)

\bibitem{Zhou2023-jk}
Zhou, K., Liu, Z., Qiao, Y., Xiang, T., Loy, C.C.: Domain generalization: A survey. IEEE Transactions on Pattern Analysis and Machine Intelligence  \textbf{45}(4),  4396--4415 (2023). \doi{10.1109/TPAMI.2022.3195549}

\bibitem{Zhuang2019MvMM}
Zhuang, X.: {Multivariate Mixture Model for Myocardial Segmentation Combining Multi-Source Images}. IEEE Transactions on Pattern Analysis and Machine Intelligence  \textbf{41}(12),  2933--2946 (2019)

\bibitem{Zhuang2019-uj}
Zhuang, X., Li, L., Payer, C., {\v S}tern, D., Urschler, M., Heinrich, M.P., Oster, J., Wang, C., Smedby, {\"O}., Bian, C., Yang, X., Heng, P.A., Mortazi, A., Bagci, U., Yang, G., Sun, C., Galisot, G., Ramel, J.Y., Brouard, T., Tong, Q., Si, W., Liao, X., Zeng, G., Shi, Z., Zheng, G., Wang, C., MacGillivray, T., Newby, D., Rhode, K., Ourselin, S., Mohiaddin, R., Keegan, J., Firmin, D., Yang, G.: {Evaluation of algorithms for {Multi-Modality} Whole Heart Segmentation: An open-access grand challenge}. Medical Image Analysis  \textbf{58},  101537 (2019). \doi{10.1016/j.media.2019.101537}

\bibitem{Zhuang2016-dz}
Zhuang, X., Shen, J.: Multi-scale patch and multi-modality atlases for whole heart segmentation of {MRI}. Medical Image Analysis  \textbf{31},  77--87 (Jul 2016). \doi{10.1016/j.media.2016.02.006}

\end{thebibliography}

\end{document}